\documentclass[letterpaper, 10 pt, conference]{ieeeconf}  

\IEEEoverridecommandlockouts                              
\usepackage[utf8]{inputenc}
\usepackage[T1]{fontenc}
\usepackage{paralist}

\usepackage{graphics} 
\usepackage{epsfig} 
\usepackage{mathptmx} 
\usepackage{mathptmx} 
\usepackage{amsmath} 
\usepackage{amssymb}  

\usepackage{multirow}
\usepackage{tabulary}
\usepackage[table]{xcolor} 
\usepackage{cite}

\definecolor{blue}{rgb}{0,0,1}
\newcommand{\blue}[1]{\textcolor{blue}{#1}}
\definecolor{green}{rgb}{0,0.5,0}
\newcommand{\green}[1]{\textcolor{green}{#1}}
\definecolor{orange}{rgb}{1,0.2,0}
\newcommand{\orange}[1]{\textcolor{orange}{#1}}

\usepackage{caption}
\let\oldtwocolumn\twocolumn
\renewcommand\twocolumn[1][]{%
    \oldtwocolumn[{#1}{
    \begin{center}
    \vskip-5.5ex
        \includegraphics[width=0.88\textwidth]{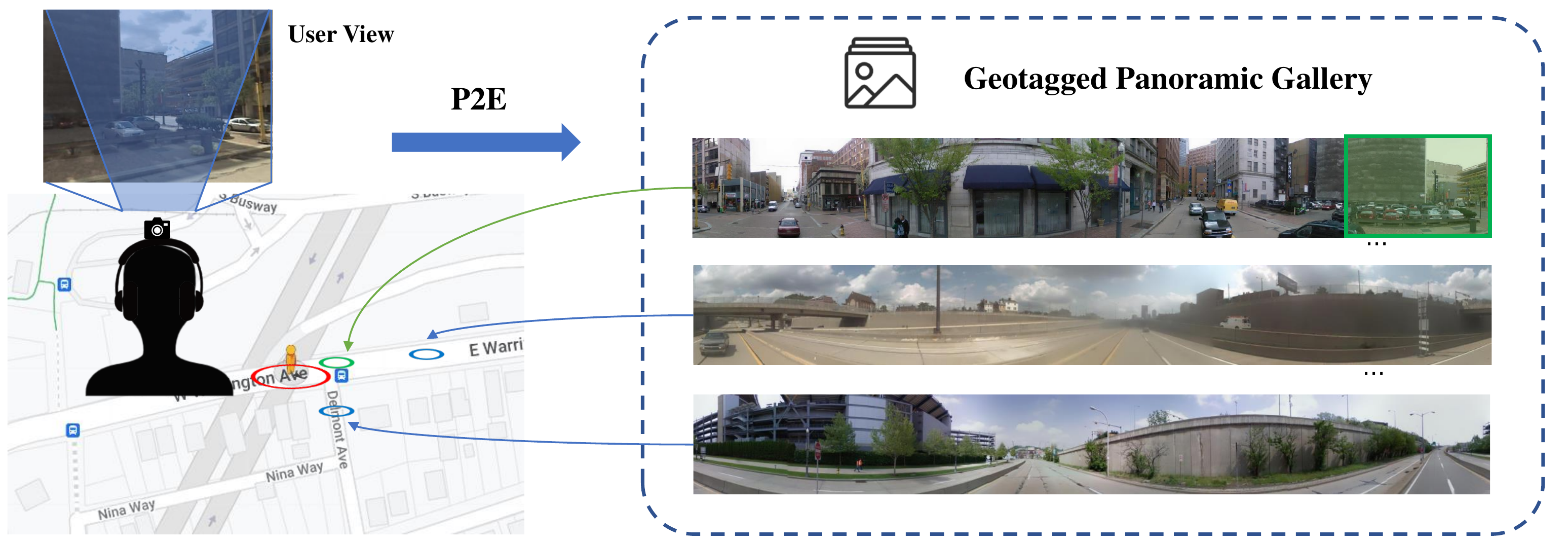}
        \end{center}
        \vskip-1em
        \captionof{figure} {
\textbf{\emph{Illustrations of perspective-to-equirectangular visual place recognition (P2E-VPR).}} Considering that users usually equip with consumer-grade pinhole cameras on mobile phones, and map providers' databases consist of panoramic images captured by spherical cameras. The left image in \blue{blue} shows the user's perspective view, while the \green{green} window on the right depicts the correct window found in the geotagged panoramic gallery.
        }
        \label{fig:teaser}
    }]
}

\makeatletter
\let\NAT@parse\undefined
\makeatother
\usepackage[colorlinks]{hyperref}
\hypersetup{colorlinks=true,linkcolor=red,citecolor=blue,urlcolor=magenta}

\definecolor{revised_color_SH}{HTML}{007FFF}

\newcommand{\etal}{\emph{et al.}}

\usepackage[capitalize]{cleveref}
\crefname{section}{Sec.}{Secs.}
\Crefname{section}{Section}{Sections}
\Crefname{table}{Table}{Tables}
\crefname{table}{Tab.}{Tabs.}

\usepackage{setspace}

\title{\LARGE \bf
PanoVPR: Towards Unified Perspective-to-Equirectangular Visual Place Recognition via Sliding Windows across the Panoramic View
}

\author{Ze Shi$^{1,*}$, Hao Shi$^{1,3,*}$, Kailun Yang$^{2}$, Zhe Yin$^{1}$, Yining Lin$^{3}$, and Kaiwei Wang$^{1}$
\thanks{This paper is supported in part by the National Natural Science Foundation of China (NSFC) under Grant No. 12174341, in part by Shanghai SUPREMIND Technology Co. Ltd, and in part by Hangzhou SurImage Technology Co. Ltd. \textit{(Corresponding authors: Kaiwei Wang; Kailun Yang.)}}
\thanks{* denotes equal contribution}
\thanks{$^{1}$Ze Shi, Hao Shi, Zhe Yin, and Kaiwei Wang are with the State Key Laboratory of Modern Optical Instrumentation, Zhejiang University, Hangzhou, China. E-mail: wangkaiwei@zju.edu.cn.}
\thanks{$^{2}$Kailun Yang is with the School of Robotics and with the National Engineering Research Center of Robot Visual Perception and Control Technology, Hunan University, Changsha, China. E-mail: kailun.yang@hnu.edu.cn.}%
\thanks{$^{3}$Hao Shi and Yining Lin are with Shanghai SUPREMIND Technology Co., Ltd, Shanghai, China. E-mail: linyining@supremind.com.}%
}

\begin{document}

\maketitle
\thispagestyle{empty}
\pagestyle{empty}

\begin{abstract}

Visual place recognition has gained significant attention in recent years as a crucial technology in autonomous driving and robotics. Currently, the two main approaches are the perspective view retrieval (P2P) paradigm and the equirectangular image retrieval (E2E) paradigm. However, it is practical and natural to assume that users only have consumer-grade pinhole cameras to obtain query perspective images and retrieve them in panoramic database images from map providers. To address this, we propose \textit{PanoVPR}, a perspective-to-equirectangular (P2E) visual place recognition framework that employs sliding windows to eliminate feature truncation caused by hard cropping. Specifically, PanoVPR slides windows over the entire equirectangular image and computes feature descriptors for each window, which are then compared to determine place similarity. Notably, our unified framework enables direct transfer of the backbone from P2P methods without any modification, supporting not only CNNs but also Transformers. To facilitate training and evaluation, we derive the Pitts250k-P2E dataset from the Pitts250k and establish YQ360, latter is the first P2E visual place recognition dataset collected by a mobile robot platform aiming to simulate real-world task scenarios better. 
Extensive experiments demonstrate that PanoVPR achieves state-of-the-art performance and obtains $3.8\%$ and $8.0\%$ performance gain on Pitts250k-P2E and YQ360 compared to the previous best method, respectively. Code and datasets will be publicly available at~\href{https://github.com/zafirshi/PanoVPR}{PanoVPR}.

\end{abstract}

\section{INTRODUCTION}

Visual place recognition refers to the process of identifying pre-stored scene locations based on image data captured by visual sensors such as cameras, which can help to obtain more accurate geographical coordinates~\cite{garg2021your}.
Related applications are widely deployed on unmanned vehicles and mobile robots~\cite{liu2018mobile,zhang2021visual}.
Generally, the visual place recognition task is redefined as a fine-grained image retrieval task that applies the image retrieval pipeline~\cite{masone_survey_2021}. 

Most image-based visual place recognition methods can be divided into two categories: \textit{perspective to perspective} (P2P)~\cite{arandjelovic2016netvlad, hausler2021patch, wang2022transvpr} or \textit{equirectangular to equirectangular}~(E2E)~\cite{fang2020cfvl,cheng2019panoramic,wang2018omnidirectional} fashion. 
However, both of the aforementioned tasks have their own limitations.
P2P methods have limitations with regard to the field of view (FoV) of the pinhole camera and the potential for inaccuracies due to different shooting orientations. The non-overlapping hard cropping method utilized in panoramic database images can also lead to mismatches for query images whose FoV stride over the cropping boundary~\cite{orhan2021efficient}. On the other hand, E2E methods are relatively complex in optical design~\cite{gao2022review} and expensive to integrate into consumer-grade hardware~\cite{broxton2019low}.

Hence, a natural and straightforward idea is to employ perspective images for retrieving panoramic database images. These perspective images can be captured by users using low-cost consumer-grade pinhole cameras in a convenient manner. Furthermore, panoramic images with location tags can be obtained from map providers, whose data are abundant and easily accessible~\cite{orhan2021efficient}.
In this way, users can obtain relatively reliable geographical coordinates in GPS-unreliable outdoor scenes through P2E (perspective-to-equirectangular) algorithms.

\begin{figure}[!t]
\renewcommand{\thefigure}{2}    %
   \centering
   \vspace{0.5em}
   \includegraphics[width=1\linewidth]{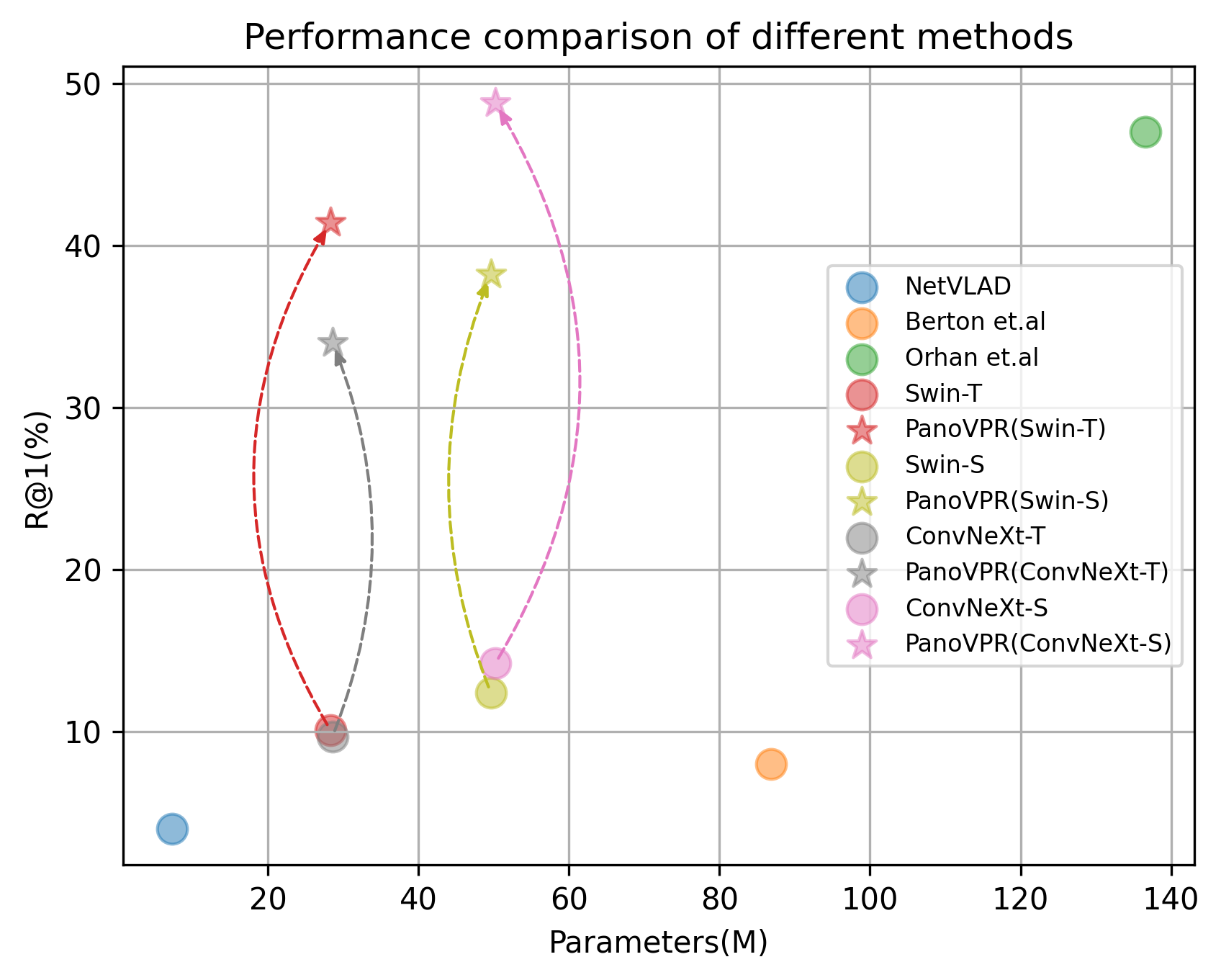}
   \vspace{-1em}
   \caption{\emph{\textbf{Performance comparison} against} NetVLAD~\cite{arandjelovic2016netvlad}, Berton~\etal~\cite{berton2022deep} \emph{and} Orhan~\etal~\cite{orhan2021efficient} \emph{with PanoVPRs on pitts250K-P2E dataset for the P2E-VPR task.}}
   \label{fig:performance_comparison}
  \vspace{-1.0em}
\end{figure}

However, the difficulty of designing a unified P2E framework is the asymmetrical data capacity and informational content between perspective and panoramic images, making it unreasonable to apply P2P or E2E frameworks directly.
In the field of visual place recognition, using a pipeline that employs consistent encoding behavior for database and query images is of vital importance. 
This is because an offline database must be constructed and online query features must be ensured. Thus, it is necessary for the \emph{query} pinhole images and the \emph{database} panoramic images to be encoded by the same encoder. 
Therefore, the utilization of a transformer-based backbone is limited in the P2E-VPR task by natural structural design, such as positional embedding~\cite{vaswani2017attention}.

To tackle these issues and enable the direct transfer of backbone networks from P2P methods without modifications, we slide a window over the equidistant cylindrical projection of the panoramic image, calculating and comparing the similarity of descriptors within the window.
Additionally, during training, we only select the window in the whole panoramic image with the closest descriptor distance in feature space relative to the query image, and then extract the descriptors of that window patch.
Window-based descriptors with the same dimension are applied to calculate the triplet loss function~\cite{balntas2016learning}, avoiding label misclassification caused by dissimilar windows in the same panoramic image.
The experiment shows that our proposed sliding window method achieves higher retrieval accuracy than resizing the panoramic image and applying the P2P framework directly.
Moreover, we integrate the sliding window strategy and proposed a unified end-to-end P2E visual place recognition framework, \textbf{PanoVPR}.
To train the proposed network on large-scale datasets, we derive a dataset by panoramic stitching database images specifically for our P2E task, called Pitts250k-P2E, based on the Pitts250k dataset~\cite{arandjelovic2016netvlad,torii2013visual}.

Further, to validate the framework's performance in real-world environments where the FoV of query images does not completely overlap with that of the panoramic database images, we collect the first P2E dataset in real-world via a mobile robot with a Panoramic Annular Lens (PAL) camera, which we refer to as YQ360.
Extensive experiments on the two datasets verify that
the sliding window strategy on panoramic images is consistently effective for various backbone networks, and a small step size with overlapping windows can achieve higher accuracy.
This conclusion holds true for different types of backbone networks used in the framework, whether a Transformer or a CNN (Fig.~\ref{fig:performance_comparison}).

In summary, our work has the following contributions:
\begin{compactitem}
    \item Proposing a method of cyclically overlapping sliding windows on panoramic images (Fig.~\ref{fig:sliding_window}), which solves the problem of completed objects being cut apart due to hard cropping of panoramic images, leading to insufficient features during retrieval. Further, our method leverages the cyclically invariant characteristics of panoramic images to solve the problem of discontinuous boundaries in unfolded panoramas (Sec.~\ref{sec:slidig_window}).

    \item Designing a sliding window-based framework named \textbf{PanoVPR} for visual place recognition from perspective to panoramic images (Fig.~\ref{fig:framework}). This unified framework can directly transfer most perspective-to-perspective methods without any modification. The feature encoding strategy using sliding windows across raw panoramic images enables compatibility with transformer-based backbones. Moreover, a window-based triplet loss function is proposed, facilitating easy training and fast convergence of the model (Sec.~\ref{sec:framework}).

    \item Proposing two datasets for the P2E task, one is the large-scale dataset derived from the Pitts250k dataset by stitching panoramas in the database Pitts250k-P2E. The other is a real-world dataset YQ360, which was collected with query images and panoramic database images having partially overlapped FoVs (Sec.~\ref{sec:dataset}).

\end{compactitem}

\section{RELATED WORK}
\noindent\textbf{Perspective Visual Place Recognition.} 
Visual place recognition from perspective to perspective involves narrow field-of-view images for both query and database.
The image data is directly acquired from a pinhole camera, such as Norland~\cite{sunderhauf2013we}, RobotCar Seasons~\cite{maddern20171}, and MSLS~\cite{warburg2020mapillary}, or by unwrapping and cropping panoramic images provided by a map provider, such as Pitts250k~\cite{arandjelovic2016netvlad,torii2013visual}, SF-XL~\cite{Berton_CVPR_2022_CosPlace}, \textit{etc.}

Initially, researchers used hand-crafted methods to detect key points in images and extract compressed representations as feature descriptors, which serve as the representation of the key points information in images, such as SIFT~\cite{ng2003sift}, SURF~\cite{bay2006surf}, and VLAD~\cite{jegou2010aggregating}.
With the popularity of deep learning, researchers found that using end-to-end neural networks, supplemented by keypoint clustering methods to extract image feature descriptors, have higher robustness than hand-crafted methods and achieve higher recall accuracy in retrieving when it comes to complex scenarios~\cite{masone2021survey, chen2022deep}.

NetVLAD~\cite{arandjelovic2016netvlad} employs a CNN backbone to extract local features from images and designs a NetVLAD layer to obtain a global feature descriptor for the image. 
Patch-NetVLAD~\cite{hausler2021patch} divides an image into patches, extracts their features using NetVLAD, and combines them with the global image descriptor to obtain a robust descriptor that considers both global and local information.
SuperPoint~\cite{detone2018superpoint} combines the interest point detection decoder and descriptor decoder in parallel at the back end of a shared CNN encoder, making it faster to infer and simpler to train. 
TransVPR~\cite{wang2022transvpr} uses a vision transformer as the backbone, to obtain the global features through multi-layer attention aggregation.

More recently, some works, such as CosPlace~\cite{Berton_CVPR_2022_CosPlace}, have converted the image retrieval problem into an image classification problem to solve the problem without expensive negative sample mining costs.
Unlike these works, we address panoramic visual place recognition which can provide a wider FoV for the image retrieval paradigm.

\noindent\textbf{Panoramic Visual Place Recognition.}
This visual place recognition method uses panoramic unwrapped images with a large field of view for both query and database images.
Wang~\etal~\cite{wang2018omnidirectional} utilize CNNs to extract features from indoor scenes and design circular padding and roll branching techniques to improve the model's performance, considering the rotational invariance characteristics of panoramic images.
Cheng~\etal~\cite{cheng2019panoramic} propose a visual localization method for challenging outdoor scenes, unwrapping panoramas into cylindrical projection as database images.
Marta~\etal~\cite{ballesta2021cnn} use CNNs to extract robust scene descriptors from omnidirectional images directly for performing hierarchical localization of a mobile robot in indoor environments.
Fang~\etal~\cite{fang2020cfvl} propose a two-stage place recognition method, which roughly filters panoramic cylindrical unwrapped database images in the first stage and re-ranks the candidate images using Geodesc~\cite{luo2018geodesc} in stage two.

Recent work on efficient retrieval of perspective images from panoramic images involves sliding windows on \emph{feature maps} generated by \emph{CNNs}~\cite{orhan2021efficient}.
CNN can handle input images of arbitrary sizes, so it is possible to encode entire panoramic images. 
However, if we use visual transformers~\cite{liu2021swin,dosovitskiy2020image} as the backbone, encoding the entire panoramic image directly would require a huge amount of GPU memory which is impractical.
Moreover, directly resizing the entire raw panorama to an appropriate size for the transformer would significantly reduce the model's accuracy, as shown in Sec.~\ref{sec:experiments}.

To increase the diversity of backbone network types in the framework and address the issue of transformer-based backbone being unable to encode entire panoramic images directly, we slide windows over the panorama and only feed the windowed portion of the panorama into the backbone.
Therefore, the backbone in the framework only computes for the windowed portion, solving the problem of encoding large-sized panoramic database images.

\section{METHODOLOGY}
\label{sec:methodology}
\begin{figure*}[!t]
\renewcommand{\thefigure}{3}    %
   \centering
   \includegraphics[width=1.0\linewidth]{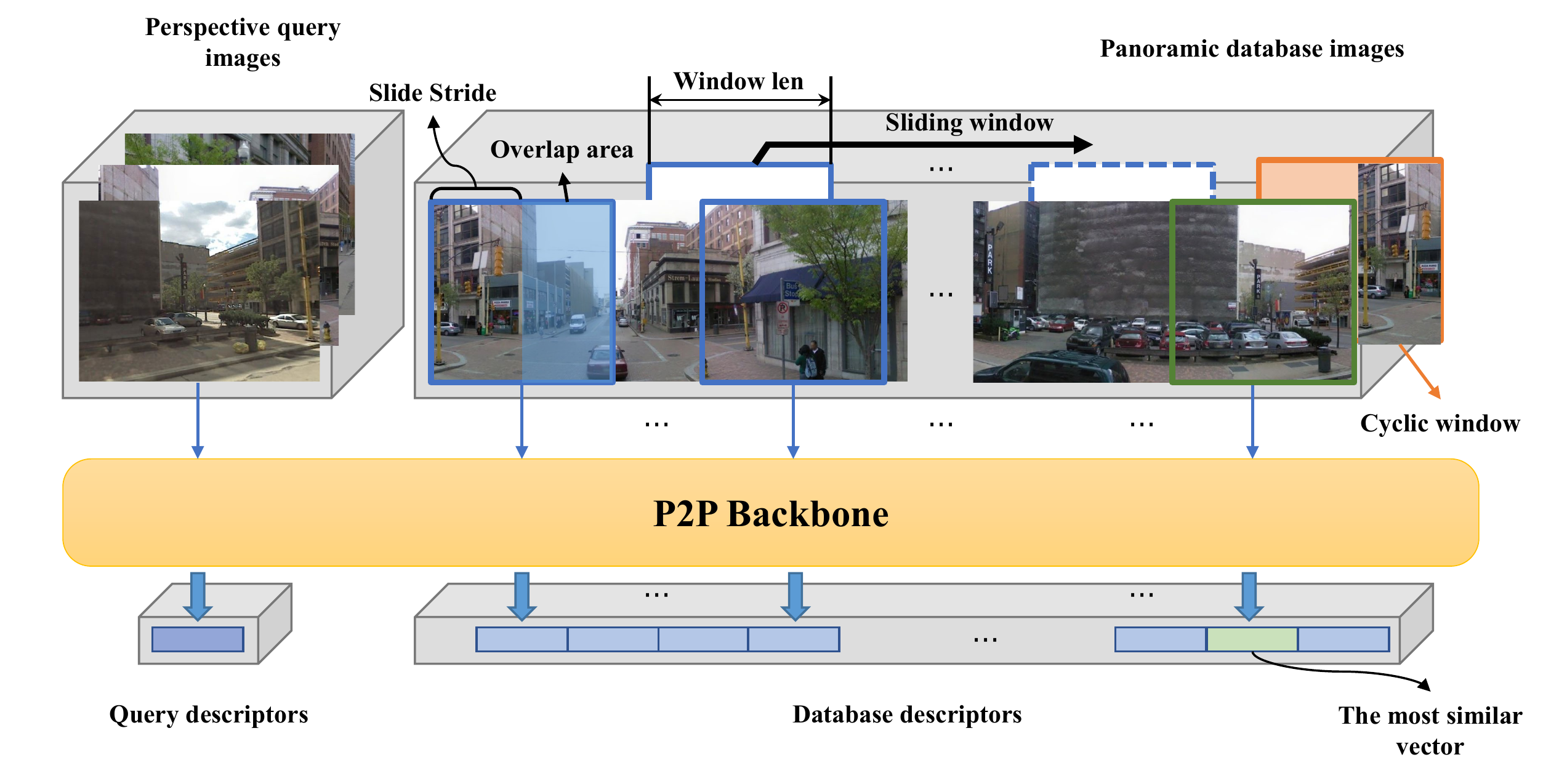}
   \vspace{-3em}
   \caption{\textbf{\emph{Illustrations of the proposed sliding window strategy.}} We slide a window over the panoramic image and encode the image in each window. The overlapping sliding window (\blue{blue} shadow in the figure) can eliminate the problem of separating entire objects due to hard image clipping. The cyclic sliding window (\orange{orange}) fully uses the cyclic invariance of the panoramic image. The \green{green} window is the one that should be retrieved correctly during the sliding search. Both the perspective and panoramic images utilize the P2P backbone to extract feature descriptors.}
   \label{fig:sliding_window}
  \vspace{-2em}
\end{figure*}

We propose a \textbf{V}isual \textbf{P}lace \textbf{R}ecognition framework for retrieving \textbf{Pano}ramic database images using perspective query images, dubbed \textbf{PanoVPR}.

Our proposed framework employs a sliding window approach and utilizes an image retrieval pipeline to provide global feature descriptors for both panoramic and perspective images.
In Sec.~\ref{sec:slidig_window}, we describe our panoramic sliding window approach, including the overlap sliding window design that eliminates the effect of seams and the cyclic sliding window strategy that utilizes the cyclic information of panoramas.
Furthermore, in Sec.~\ref{sec:framework}, we detail our unified P2E visual place recognition framework based on the sliding window approach and elaborate the workflow of PanoVPR in the training and interface process. 
Besides, we also explain the window-based modification to the loss function, which makes the training process more reasonable and stable. At the end of the section, we provided a plain and simple explanation for the evaluation metric of the framework.

\subsection{Sliding Window}
\label{sec:slidig_window}

As there are discrepancies in input sizes between perspective and panoramic images, using the transformer-based P2P backbone directly is not feasible due to the need for homogeneous encoding behavior like positional embedding.
To address this challenge and create a unified framework that can leverage the P2P visual place recognition backbone directly, we employ a sliding window approach on the panoramic database images. 
This approach narrows the model's observation range of the large field of view panoramas and transforms the problem into a comparison-and-retrieval process within the window.

As shown in Fig.~\ref{fig:sliding_window},  we utilize a sliding window approach on panoramic images, dividing them into multiple sub-images. 
Each sub-image between the same panoramic image is independent and processed by the same encoder to extract features.
It could be intuitively understood that the \emph{window} refers to the area of interest during the sliding retrieval phase, and the sliding operation represents the process of horizontal interested region change in the scene captured by this panoramic image.
Specifically, the \blue{blue} region depicted in Fig.~\ref{fig:sliding_window} corresponds to an overlapping sliding window applied to the panoramic image. The shaded portion within the blue region represents the overlapping segment of the sliding window. 
To emphasize the crucial elements, we only explicitly illustrate part of all blue windows.
The blue window accompanied by the rightward arrow, concealed behind the panorama, stands for the dynamic process of the overlapping sliding window as it traverses the panorama from left to right.
The \orange{orange} window in the figure represents the cyclic sliding window. 
Since the image of the panorama has continuous content at both the left and right borders, the section beyond the right border of the panorama is complemented by the left border, thereby leveraging the cyclical invariance property of the panoramic image.
The \green{green} window is the one that should be retrieved correctly during the sliding search.
It can be observed that the contents of the green window are similar in appearance to the first perspective query image by zooming in.

There are two strategies for sliding window search based on different strides: overlapping sliding windows and non-overlapping sliding windows.
In contrast to the latter, the former contains a section of images from the previous sliding window within each sliding window, which mitigates the risk of continuous feature loss at the seam. 
This, in turn, leads to enhanced accuracy in feature extraction and matching, detailed in Tab.~\ref{tab:panoVPR_pitts}. 
Meanwhile, overlapping sliding windows can also increase the number of samples and improve the robustness of the model.
Furthermore, the model's metric accuracy can be further enhanced by applying a cyclic sliding window approach, and the comparative experiment results are also available in Tab.~\ref{tab:panoVPR_pitts}.

It should be emphasized that the length of each sliding window on the panoramic image is equal to the horizontal size of the perspective query image, which is set uniformly.
This enables the utilization of the same encoder for extracting descriptors from both the query and the windowed database image, potentially satisfying the requirement for consistent query-database encoding behavior in VPR.

The entire panoramic descriptor is obtained by concatenating the sub-window vectors.  
To obtain the correct window on panoramas, we slide the query descriptor over the panoramic descriptor with a step equal to the query descriptor's length.
As shown in Fig.~\ref{fig:sliding_window}, the window with the highest similarity score between the query and sub-window database descriptor is the correct window in the panoramic image that matches the query image (window and descriptor colored in \green{green}).
Then the panorama to which it belongs provides the geographic location for the query image.

\begin{figure*}[!t]
\renewcommand{\thefigure}{4}    %
   \centering
   \vspace{-3.5em}
   \includegraphics[width=1.0\linewidth]{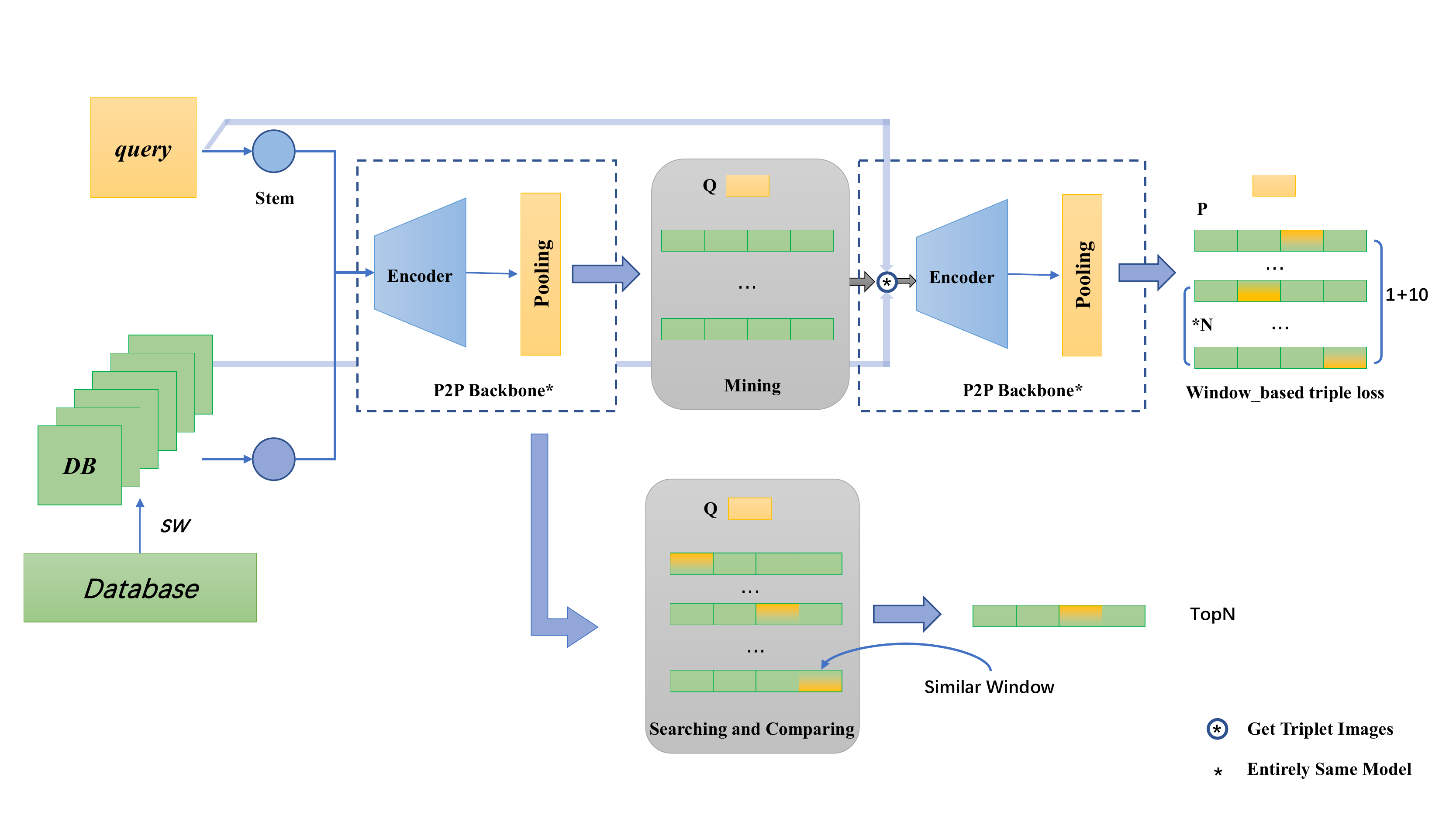}
   \vspace{-3em}
   \caption{\textbf{\emph{Illustrations of the proposed PanoVPR framework.}} During training, perspective query images and panoramic database images are fed into a shared encoder to extract features, which are then used to obtain triplet image pairs using hard positive and negative sample mining. The triplet image pairs are encoded and the feature distances are computed using a window-based triplet loss. During testing, the feature descriptors of the panoramic database images are first extracted offline, and then the perspective query images are encoded one after another. The query descriptors are compared with those of the panoramic database descriptors to obtain the Top-N predictions with the window similarity.}
   \label{fig:framework}
  \vspace{-1.0em}
\end{figure*}

\subsection{Framework}
\label{sec:framework}

\noindent\textbf{PanoVPR Framework}. 
As shown in Fig.\ref{fig:framework}, our PanoVPR framework mainly follows an image retrieval pipeline~\cite{lowry2016visual}.
In this pipeline, the training stage is completed online, where both the query images and the database images are fed into the network synchronously. While the testing stage is completed offline, where the model first extracts features from the database images and stores them, then processes the query images in streaming, with the two operations being asynchronous.
The inconsistency between the training and testing logic requires that the query images and the database images share the same feature encoding network.

During training, the model first infers all query and database images to obtain feature descriptors for all images. Then the hard positive and hard negative samples mining is conducted on the database images with the aid of GPS ground truth.
Positive samples are defined as database images that are similar to the query image both in terms of geographical coordinates and feature space, while negative samples are database images that are only similar in feature space but are far away in geographical coordinates. 
After the hard mining process, we obtain many triplets image pairs containing a query image, a hard positive sample, and multiple hard negative samples:
\begin{equation}
    (q,db) \longrightarrow (q, p, *n),    
\end{equation}
where $q$ and $db$ refer to perspective query images and panoramic database images respectively, $p$ refers to a hard positive panoramic image, and $*n$ denotes a collection of hard negative panoramic images.
The obtained triplets are fed into the backbone for subsequent training. 

During the testing phase, we first encode the panoramic database images using the backbone and store them as descriptors. 
Next, for each perspective query image, we encode it using the same backbone to generate a query feature descriptor. We slide this descriptor on all the database descriptors and calculate the similarity score for each local window. 
The panoramic database image with the highest similarity score within the searched window is selected as the top-$1$ predicted image, and this process is repeated for all descriptor windows. 
Using this search method, we reorder all database images according to their local window similarity. Finally, we select the top-$N$ images to obtain the top-$N$ predictions for each perspective query image.

\noindent\textbf{Window-based Triplet Loss}. 
Since the PanoVPR framework extracts perspective query images and panoramic database images with different descriptor lengths, we design window-based triplet loss based on the original triplet loss~\cite{balntas2016learning}.
The original triplet loss function is defined as:
\begin{equation}
    loss(q,p,*n)=\sum_{i}^{}  max\{d(q,p)-d(q,n_i)+m, 0\},
\end{equation}
where $m$ is the margin parameter, and $d(x, y)$ is the distance between the feature vectors $x$ and $y$, defined as follows:
\begin{equation}
    d(x, y)=||x-y||_p.
\end{equation}

The distance mentioned above can serve as a measure of the similarity score between query and database descriptors in feature space. 
The smaller the distance value, the higher the similarity score.
We only calculate the windowed database descriptor that exhibits the highest similarity to the query descriptor in the feature space.
Therefore, the above equation is modified below:
\begin{equation}
    d(x, y)=||x-y_w||_p,
\end{equation}
where $y_w$ represents the sliding window component of vector $y$ that has the most similar to $x$, and it can be mathematically represented as:
\begin{equation}
    y_w=argmin_{win}(||x-y[win]||_p).
\end{equation}

\noindent\textbf{Evaluation Metric.}
Following ~\cite{lowry2016visual}, we use the recall metric to estimate the model's performance. The recall metric is defined as follows: given a query image, we search for the database images and predict many retrieval candidates. If the ground truth is included in the top-$N$ predictions, then the prediction for this query image is correct. The percentage of correct predictions for all query images is calculated and denoted as Recall@N.

\section{EXPERIMENTS}
\label{sec:experiments}
\subsection{Datasets}
\label{sec:dataset}
Currently, there are few readily available datasets for the P2E place recognition task.
To conduct experiments and facilitate future research, we propose two datasets specifically for the P2E task, including \emph{Pitts250K-P2E} and \emph{YQ360} datasets, respectively.
The scale and division of the two datasets are shown in Table \ref{tab:dataset}.

\noindent\textbf{Pitts250k-P2E.}
Pitts250k-P2E is derived from the original Pitts250k dataset~\cite{arandjelovic2016netvlad,torii2013visual}.
In the original dataset, each location's panoramas were captured in $2$ yaw directions, we select $12$ images with lower yaw angles for each location as they contained more street-view information instead of irrelevant background elements such as the sky and less distortion. 
These images are then stitched together to create one panoramic image~\cite{brown2007automatic} for each location.
As for the query image, we select $12$ images with lower yaw angles that have overlapping fields of view (FoV) with the panoramic image but are captured at different times.

\begin{figure}[!t]
\renewcommand{\thefigure}{5}    %
   \centering
   \includegraphics[width=0.95\linewidth]{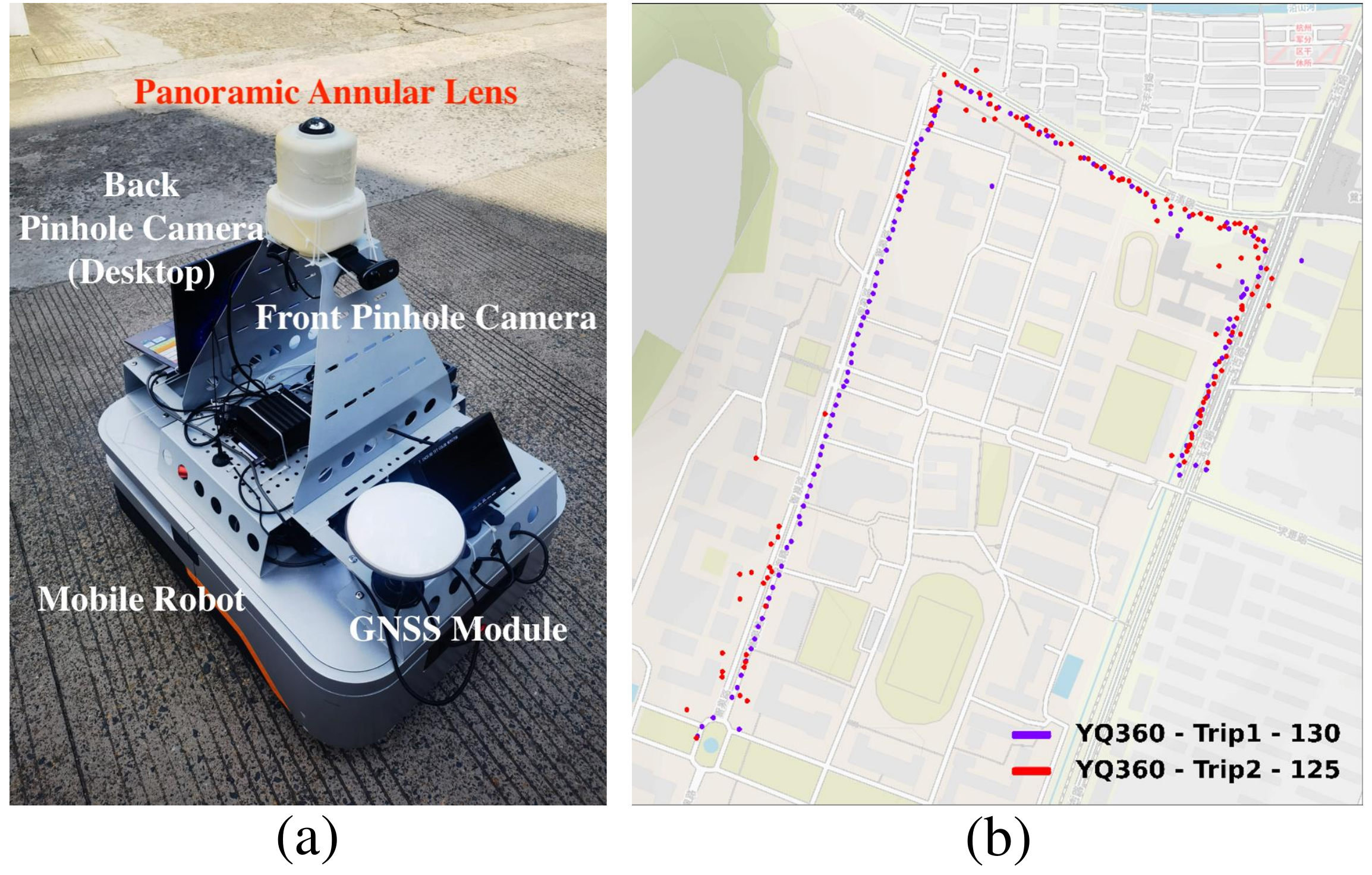}
   \vspace{-0.5em}
   \caption{\textbf{\emph{YQ360 Datasets}}. (a) The data collection devices, include front and rear pinhole cameras, panoramic annular cameras, and GNSS. (b) The visualization of the dataset.}
   \label{fig:YQ360}
  \vspace{-1.5em}
\end{figure}

\noindent\textbf{YQ360.}
The YQ360 dataset is collected using a mobile vehicle in Fig.~\ref{fig:YQ360}(a), which
is equipped with a Panoramic Annular Lens (PAL) camera on the top.
Meanwhile, a network camera and a laptop's built-in camera are mounted on the front and rear of the car respectively, serving as the front and rear pinhole cameras.
The GNSS sensor is also placed on the vehicle to obtain the GPS coordinates of the acquisition location.
We capture a panoramic image and two narrow-field images (one from the front and another from the rear) approximately every $5$ meters, tagged with the GNSS output of the location. 
To simulate a more realistic testing scenario, the vertical fields of view of the perspective and panoramic images are not completely overlapped, which is also an important characteristic of YQ360. 
The collection routine covers the same path twice as shown in Fig.~\ref{fig:YQ360}(b).
It includes a boulevard with heavily repeated textures, as well as open street views that have distinctive features.
The scale of the YQ360 dataset is currently constrained owing to the discrete data acquisition process from various sensors in different outdoor locations, which is labor-intensive.

\subsection{Implementation Details}
Our framework is implemented using PyTorch and the experiments are conducted on a machine with $4$ NVIDIA 3090 graphics cards, 128 GB RAM, and 88 CPU cores.
During training, each panoramic image is transformed into multiple sub-images using our sliding window method. 
GPU memory consumption increases with the number of sub-images, the training batch size is set to $2$ considering that.
We also set the margin parameters $m$ in triplet loss to $0.1$. 
Moreover, when testing on the YQ360 dataset, we consider the variations in the overlap of the field of view between the query and database images from different datasets.
Specifically, we fine-tune our model, pre-trained on pitts250k-P2E, using the training and validation sets of the YQ360 dataset. The fine-tuned model is then evaluated on the test set.

As conventional practice in ~\cite{arandjelovic2016netvlad}, during the mining period, $1$ hard positive sample and $10$ hard negative samples are mined for each query image.
The geographical coordinate labels are clustered using KNN~\cite{taunk2019brief} method, with a clustering threshold of $10$ meters during training and $25$ meters during testing.
In~\cite{berton2022deep}, it has been shown that the \emph{partial} mining method can achieve accuracy similar to the \emph{full} mining approach while reducing memory usage. 
Therefore, we use the \emph{partial} mining method in the following experiments. 
Moreover, the back-end of the framework utilizes the GeM pooling layer~\cite{radenovic2018fine}. 

\subsection{Results}
\noindent\textbf{Sliding Window Validation.}
To verify the effectiveness of our proposed sliding window method on panoramic images, we conduct the following experiments on the Pitts250k-P2E dataset and show the results in Table \ref{tab:panoVPR_pitts}. 
We select the lightweight Swin-Tiny~\cite{liu2021swin} as our backbone and directly resize panoramas to meet the requirements for input image size as our baseline.
$\times N$ indicates that the panoramic image is divided into $N$ equal parts, and the sliding window's stride selects the length value of one of them ($1/N$ of the width of the panorama). 
As the sliding window size is predetermined (discussed in Sec.~\ref{sec:slidig_window}), larger values of $N$ correspond to smaller strides and more refined sliding windows.

\begin{table}[!t]
\renewcommand{\thetable}{1}
    \begin{center}
        \caption{The training, validation, and testing splits.}
        \label{tab:dataset}
\resizebox{1.0\columnwidth}{!}{
\setlength{\tabcolsep}{1mm}{ 
\begin{tabular}{l|ccccc}
 \hline
    \textbf{Dataset} & \textbf{Type} & \textbf{Train} & \textbf{Validation} & \textbf{Test} & \textbf{Total} \\
 \hline\hline
 
    \multirow{2}{*}{Pitts250k-P2E} & Query & 2,940 & 3,804 & 4,140 & 10,884\\
     & Database & 2,466 & 2,116 & 2,158 & 6,720\\
     
 \hline 
 
    \multirow{2}{*}{YQ360} & Query & 182 & 78 & 250 & 310\\
     & Database & 91 & 39 & 125 & 255\\

 \hline
\end{tabular}
}
}

        \vspace{-1em}
    \end{center}
\end{table}

\begin{table}[!t]
\renewcommand{\thetable}{2}
    \begin{center}
        \caption{Effectiveness of PanoVPR on Pitts250k-P2E.}
        \label{tab:panoVPR_pitts}
        \resizebox{1.0\columnwidth}{!}
{
\setlength{\tabcolsep}{1mm}{ 
\begin{tabular}{l|cc|cccccc}
\hline
\textbf{Dataset}   & \multicolumn{2}{c|}{\textbf{Config}} & \multicolumn{5}{c}{\textbf{Pitts250k-P2E}}                        \\

\hline

Method   &overlap & cycle  &R@1 & R@5 & R@10 & R@20 & Diff.@1     \\ 

\hline     
\hline  

 SwinT~\cite{liu2021swin}                   & - & -      & 10.1 & 26.3 & 36.0 & 47.4 & -  \\
\hline 
 PanoVPR (SwinT)$_{{\times}8}$            &\texttimes &\texttimes                       & 22.0 & 42.2 & 51.8 & 62.1 & \blue{+ 11.9} \\
 PanoVPR (SwinT)\textsuperscript{*}$_{{\times}16}$          &\checkmark(50\%) &\texttimes        & 32.4 & 55.8 & 65.9 & 74.1 & \blue{+ 22.3} \\
 PanoVPR (SwinT)$_{{\times}16}$           &\checkmark(50\%) &\checkmark      & 33.6 & 56.7 & 66.4 & 74.4 & \blue{+ 23.5} \\
 PanoVPR (SwinT)$_{{\times}24}$           &\checkmark(66\%) &\checkmark      & 37.8 & 59.7 & 68.5 & 76.4 & \blue{+ 27.7} \\
 PanoVPR (SwinT)$_{{\times}32}$           &\checkmark(75\%) &\checkmark     & \textbf{41.4} & \textbf{64.0} & \textbf{72.2} & \textbf{79.4} & \blue{+ 31.3} \\
\hline
\end{tabular}
}
}

        \vspace{-1em}
    \end{center}
\end{table}

\begin{table}[!t]
\renewcommand{\thetable}{3}
    \begin{center}
        \caption{Quantitative comparison on the YQ360 dataset.}
        \label{tab:panoVPR_yq}
        \resizebox{1.0\columnwidth}{!}{
\setlength{\tabcolsep}{1mm}{ 
\begin{tabular}{lc|ccccc}
\hline
 {Dataset}   & & \multicolumn{4}{c}{ {YQ360}}                                                \\

\hline

Method & \#Params.   &R@1 & R@5 & R@10 & R@20    \\ 

\hline     
\hline  

 NetVLAD~\cite{arandjelovic2016netvlad} & 7.23  & 35.2 & 66.8 & 80.0 & 90.8\\
 Berton~\etal~\cite{berton2022deep} & 86.86      & 40.4 & 74.8 & 88.4 & 95.8\\
 Orhan~\etal~\cite{orhan2021efficient}& 136.62   & 47.6 & 79.2 & 88.4 & 95.2\\

 \hline 
 Swin-T~\cite{liu2021swin}  & 28.29                                 & 27.4 & 63.5 & 72.3 & 86.1\\
 PanoVPR (SwinT)$_{{\times}8}$ & 28.29                                 & 30.8 & 69.6 & 81.6 & 92.4\\
 PanoVPR (SwinT)$_{{\times}16}$ & 28.29                                & 43.2 & 82.4 & 90.8 & 96.4\\
 PanoVPR (SwinT)$_{{\times}24}$ & 28.29                                & 48.4 & \underline{88.0} &  \textbf{94.8} &  \underline{97.6}\\

 \hline
 Swin-S~\cite{liu2021swin} & 49.61                                 & 30.1 & 66.4 & 77.2 & 89.6\\
 PanoVPR (SwinS)$_{{\times}8}$ & 49.61                                 & 32.6 & 70.0 & 85.6 & 93.2\\
 PanoVPR (SwinS)$_{{\times}16}$ & 49.61                               & 36.8 & 78.2 & 86.2 & 95.8\\
 PanoVPR (SwinS)$_{{\times}24}$ & 49.61                               &  {42.4} &  {82.7} &  {92.4} &  {96.4}\\

 \hline
 ConvNeXt-T~\cite{liu2022convnet} & 28.59                                 & 35.7& 75.2 & 86.4 & 93.6\\
 PanoVPR (ConvNeXtT)$_{{\times}8}$ & 28.59                             & 39.6 & 76.8 & 87.6 & 94.0\\
 PanoVPR (ConvNeXtT)$_{{\times}16}$ & 28.59                             & 45.6 & 84.0 & 90.0 & 96.0\\
 PanoVPR (ConvNeXtT)$_{{\times}24}$ & 28.59                             &  \underline{49.2} &  {86.0} &  {90.8} &  {96.8}\\

 \hline
 ConvNeXt-S~\cite{liu2022convnet} & 50.22                                 & 37.0 & 72.5 & 83.9 & 94.1\\
 PanoVPR (ConvNeXtS)$_{{\times}8}$ & 50.22                             & 41.2 & 75.2 & 85.2 & 96.4\\
 PanoVPR (ConvNeXtS)$_{{\times}16}$ & 50.22                             & 46.0 & 83.2 & 92.4 & 97.2\\
 PanoVPR (ConvNeXtS)$_{{\times}24}$ & 50.22                             &  \textbf{51.4} &  \textbf{89.2} &  \underline{93.1} &  \textbf{98.4}\\
\hline

\end{tabular}
}
}
\vspace{-1em}

        \vspace{-2em}
    \end{center}
\end{table}

\begin{figure*}[!t]
\renewcommand{\thefigure}{6}    %
   \centering
   \includegraphics[width=1.0\linewidth]{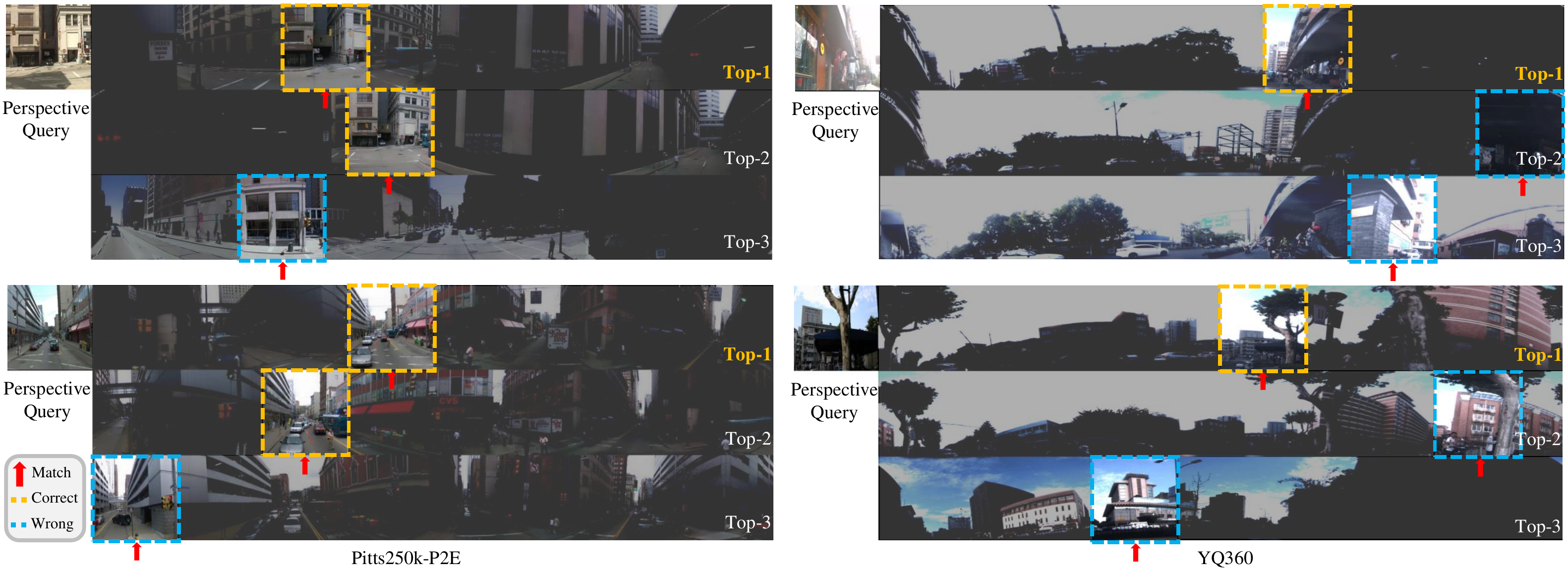}
   \vspace{-2em}
  \caption{\emph{\textbf{Visualization results on the Pitts250k-P2E and YQ360 test set using the proposed PanoVPR (SwinT) framework.} We highlight the predicted key windows in the Top3 matched database panoramic images, where the most important Top1 has achieved precise matching and positioning. PanoVPR can accurately predict the perspective query's geolocalization and focus on relevant regions of the panoramas.}}
   \label{fig:quality_pitts250k}
  \vspace{-2em}
\end{figure*}

In our experimental setup, we configured the panoramic image width to be 8 times the length of the window. Therefore, the symbol $\times 8$ indicates that the sliding window is non-overlapping, non-cycling, and equal to the stride.
We experiment with whether use circular sliding windows while maintaining the same window overlap. A comparison of the results between the third and fourth rows of Tab.~\ref{tab:panoVPR_pitts} reveals that utilizing overlapping sliding windows can lead to a $1.2\%$ increase in R@1, as well as a general improvement in other metrics.
Furthermore, we experiment with various sliding window strides \{$\times8$, $\times16$, $\times24$, $\times32$\} and find that as the stride decreases, the model's recall precision improves (from R@1 $22.0$ to $41.4$ when striding from $\times8$ to $\times32$). 

Comparisons between our method PanoVPR (SwinT) and baseline show that directly resizing panoramic images to the appropriate size for the encoder greatly reduces the model's accuracy, only reaching $10.1$ R@1 on the Pitts250k-P2E dataset, which is $31.3$ points lower than our method \emph{PanoVPR (SwinT)$_{{\times}24}$}.
We hypothesize that the reason for this is that resizing panoramic images causes a significant loss of information, resulting in distorted global features that are no longer accurate or reliable after encoding. 
However, with our proposed approach, the encoder receives panoramic images through sliding image windows, which preserves their original integrity and avoids distortion before encoding.
Therefore, our proposed method significantly improves the accuracy compared to the baseline.

\noindent\textbf{Comparison against State-of-the-Art.}
We conduct the quantitative comparison on two datasets and show the results in Tab.~\ref{tab:panoVPR_yq} and Tab.~\ref{tab:abl_pitts250k}, respectively.
In NetVLAD~\cite{arandjelovic2016netvlad}, the backbone is selected as ResNet50, and other settings remain unchanged.
In~\cite{berton2022deep}, the backbone is selected as ViT, and the clustering back-end uses GeM~\cite{radenovic2018fine}. The best result is \textbf{bolded} and the second best result is \underline{underlined}.

As shown in Tab.~\ref{tab:panoVPR_yq}, when employing Swin-Tiny as PanoVPR's backbone, it achieves a $37.5\%$ increase in R@1 accuracy compared to the popular CNN method~\cite{arandjelovic2016netvlad} ($48.4$ \textit{vs.} $35.2$). Besides, PanoVPR improves R@1 by $19.8\%$ ($48.4$ \textit{vs.} $40.4$) compared to the ViT method~\cite{berton2022deep} while reducing the model parameter amount by $67.4\%$ ($28.29M$ \textit{vs.} $86.86M$).
Furthermore, PanoVPR (ConvNeXtS) can achieve $7.98\%$ ($51.4$ \textit{vs.} $47.6$) improvement over the previous best P2E method~\cite{orhan2021efficient} in R@1 on the newly proposed YQ360 dataset, verifying our proposed method in real scenes. 
As shown in Tab.~\ref{tab:abl_pitts250k}, PanoVPR also achieves 48.8 R@1 on the derived pitts250k-P2E dataset, surpassing the previous best method by $3.8\%$ ($48.8$ \textit{vs.} $47.0$) while reducing $63.2\%$ parameters ($50.22M$ \textit{vs.} $136.62M$).
We found that PanoVPR does not require a complex model with a large number of parameters. 
In summary,
our proposed PanoVPR framework is robust and powerful that demonstrates excellent compatibility with the backbone, as well as high accuracy in localization.
From the analysis above, it is evident that our proposed sliding window approach leads to a higher recall when perspective query images retrieve panoramic images with incomplete overlapping FoV in real scenarios.

Qualitative results are shown in Fig.~\ref{fig:quality_pitts250k}, which confirm PanoVPR continues to provide accurate retrieval and localization capabilities in response to illumination, viewpoint changes, and image distortion, while also providing stable generalization across different datasets.

\begin{table}[!t]
\renewcommand{\thetable}{4}
    \begin{center}
        \caption{Quantitative comparison on Pitts250k-P2E.}
        \label{tab:abl_pitts250k}
        \resizebox{1.0\columnwidth}{!}{
\setlength{\tabcolsep}{1mm}{ 
\begin{tabular}{l|cccccc}
\hline

\textbf{Method}    & \textbf{R@1} & \textbf{R@5} & \textbf{R@10} & \textbf{R@20} & \textbf{Diff.@1}   \\ 

\hline     
\hline  
 NetVLAD~\cite{arandjelovic2016netvlad} & 4.0 & 12.4 & 20.0 & 30.2 & -\\
 Berton~\etal~\cite{berton2022deep} & 8.0 & 23.0 & 33.0 & 44.7 & -\\
 Orhan~\etal~\cite{orhan2021efficient} & \underline{47.0} & \underline{66.4} & \underline{73.6} & 80.3 & -\\

 \hline

 Swin-T~\cite{liu2021swin}                              & 10.1 & 26.3 & 36.0 & 47.4 & -\\
 PanoVPR (SwinT)$_{{\times}8}$                                   & 22.0 & 42.2 & 51.8 & 62.1 & \blue{+ 11.9} \\
 PanoVPR (SwinT)$_{{\times}16}$                                  & 33.6 & 56.7 & 66.4 & 74.4 & \blue{+ 23.5}\\
 PanoVPR (SwinT)$_{{\times}24}$                                  & 37.8 & 59.7 & 68.5 & 76.4 & \blue{+ 27.7}\\
 PanoVPR (SwinT)$_{{\times}32}$                                  & 41.4 & 64.0 & 72.2 & 79.4 & \blue{+ 31.3} \\
 
 \hline
 
  Swin-S~\cite{liu2021swin}                                   & 12.4 & 30.7 & 43.8 & 58.6 & -\\
 PanoVPR (SwinS)$_{{\times}8}$                                   & 25.8 & 49.6 & 59.6 & 69.2 & \blue{+ 13.4}\\
 PanoVPR (SwinS)$_{{\times}16}$                                  & 29.6 & 54.4 & 63.3 & 72.2 & \blue{+ 17.2}\\
 PanoVPR (SwinS)$_{{\times}24}$                                  & 34.6 & 58.4 & 67.6 & 75.4 & \blue{+ 22.2}\\
 PanoVPR (SwinS)$_{{\times}32}$                                  & 38.2 & 60.8 & 69.1 & 77.5 & \blue{+ 25.8}\\
 
 \hline
 
 ConvNeXt-T~\cite{liu2022convnet}                                   & 9.7 & 19.6 & 32.4 & 41.3 & -\\
 PanoVPR (ConvNeXtT)$_{{\times}8}$                                   & 17.5 & 36.4 & 46.9 & 58.0 & \blue{+ 7.8}\\
 PanoVPR (ConvNeXtT)$_{{\times}16}$                                  & 29.0 & 51.7 & 62.1 & 72.3 & \blue{+ 19.3}\\
 PanoVPR (ConvNeXtT)$_{{\times}24}$                                  & 34.0 & 56.4 & 66.0 & 75.1 & \blue{+ 24.3}\\

\hline

 ConvNeXt-S~\cite{liu2022convnet}                                   & 14.2 & 28.8 & 39.6 & 48.8 & -\\
 PanoVPR (ConvNeXtS)$_{{\times}8}$                                   & 30.9 & 53.9 & 64.3 & 73.9 & \blue{+ 16.7}\\
 PanoVPR (ConvNeXtS)$_{{\times}16}$                                  & 40.3 & 63.0 & 72.1 & \underline{80.5} & \blue{+ 26.1}\\
 PanoVPR (ConvNeXtS)$_{{\times}24}$                                  & \textbf{48.8} & \textbf{73.8} & \textbf{82.4} & \textbf{89.4} & \blue{+ 34.6}\\

 \hline
 
\end{tabular}
}
}
        \vspace{-2em}
    \end{center}
\end{table}

\noindent\textbf{Ablation Studies.}
To investigate the impact of different backbone networks and the sliding window stride on our framework, we conduct the following ablations.

We select a CNN backbone ConvNeXt~\cite{liu2022convnet}, and a Transformer backbone Swin Transformer~\cite{liu2021swin}. The results are shown in Tab.~\ref{tab:abl_pitts250k}.
The result demonstrates that $\times24$ sliding window stride achieves at least $24.3\%$ R@1 metric improvement over the corresponding baseline model ($34.0$ \textit{vs.} $9.7$). We also found that compared to SwinS, SwinT generalizes better on both datasets.
The conclusion that a small stride overlapping sliding window can achieve higher recall also holds true when the PanoVPR backbone changes, whether CNN or ViT architecture. Thus, PanoVPR can easily benefit from progress in classical P2P VPR methods.

\section{CONCLUSIONS}
We propose a new framework \textbf{PanoVPR}, based on sliding window to solve the problem of visual place recognition from perspective to equirectangular images. 
Our framework obtains the perspective query image's geographical location by calculating and matching the descriptors generated from the sliding window. 
The benefit of our framework lies in its ability to avoid hard cropping and allow direct transferring of the perspective-to-perspective visual place recognition backbones without any modification, supporting not only CNNs but also Transformers.
To tackle the problem of mismatched descriptor lengths between query and database, which prevents calculating triplet loss, we propose a window-based triplet loss function to train the model more effectively. 
We also derive a new large-scale dataset \emph{Pitts250k-P2E}, with geographical location tags for the P2E task. Furthermore, to better simulate real-world scenarios, we collected another P2E dataset \emph{YQ360}, in which the perspective query images and panoramic database images have non-overlapping FoV. 
Experimental results show that our proposed PanoVPR framework with a lightweight backbone achieves higher recall than our baseline and STATE-OF-THEART approaches.

\bibliographystyle{IEEEtran}
\bibliography{IEEEabrv,reference}

\end{document}